\documentclass{article} 
\usepackage{iclr2024_conference,times}

\usepackage{amsmath,amsfonts,bm}

\def\eqref#1{equation~\ref{#1}}

\def\1{\bm{1}}

\DeclareMathAlphabet{\mathsfit}{\encodingdefault}{\sfdefault}{m}{sl}
\SetMathAlphabet{\mathsfit}{bold}{\encodingdefault}{\sfdefault}{bx}{n}

\usepackage[colorlinks,citecolor=blue]{hyperref}
\usepackage{url}
\usepackage{graphicx}
\usepackage{subcaption}
\usepackage{tabularray}
\usepackage{tabu}
\usepackage{multirow}
\usepackage{multicol} 

\title{360Zhinao Technical Report}

\author{
360Zhinao Team\\
\texttt{g-zhinao-opensource@360.cn}
}

\iclrfinalcopy 
\begin{document}

\maketitle

\begin{abstract}
We present 360Zhinao models with 7B parameter size and context lengths spanning 4K, 32K and 360K, all available at \url{https://github.com/Qihoo360/360zhinao}.
For rapid development in pretraining, we establish a stable and sensitive ablation environment to evaluate and compare experiment runs with minimal model size.
Under such guidance, we perfect our data cleaning and composition strategies to pretrain $\texttt{360Zhinao-7B-Base}$ on 3.4T tokens.
We also mainly emphasize data during alignment, where we strive to balance quantity and quality with filtering and reformatting. With tailored data, 360Zhinao-7B's context window is easily extended to 32K and 360K. RMs and RLHF are trained following SFT and credibly applied to specific tasks.
All together these contributions lead to 360Zhinao-7B's competitive performance among models of similar size.

\end{abstract}

\section{Introduction}
In recent years, the field of natural language processing (NLP) has witnessed a profound transformation, fueled by the advent of large language models (LLMs) \citep{bubeck2023sparks,touvron2023llama, achiam2023gpt}, which have emerged as a cornerstone to revolutionize the way we understand and generate human language. LLMs represent a new paradigm in artificial intelligence (AI) research, characterized by their immense scale, complexity, and versatility \citep{zhao2023survey}. Those models, typically built upon advanced neural network architectures like Transformers, are trained on vast amounts of text data, encompassing billions or even trillions of words. The extensive training endows LLMs with a deep understanding of linguistic structures, nuances, and context, enabling them to generate human-like text and perform a myriad of NLP tasks with unprecedented accuracy and fluency \citep{yang2024harnessing}.

Despite the impressive capabilities of LLMs, training an LLM from scratch still struggles with several challenges. The training journey can be divided into two stages: the pretraining stage and the alignment stage \citep{zhang2023instruction}. The pretraining stage involves the model learning on large-scale textual data to build its foundational knowledge and language comprehension. However, two obstacles stick out in the pretraining stage \citep{zhao2023survey}. First, refining the training corpus to enhance the base model's performance is paramount given the enormity of pretraining data. While extensive research has delved into data cleaning and sampling methodologies \citep{soldaini2024dolma, penedo2023refinedweb, wenzek2019ccnet, gunasekar2023textbooks}, the sheer scale and intricacy of pretraining datasets still leave ample room for elevating informational density and efficiency. Second, establishing a stable and sensitive ablation environment for accurately assessing data strategies poses another challenge \citep{chang2024survey,zhou2023don}. The widely used Opencompass \citep{contributors2023opencompass} framework proves inconvenient for data strategy explorations. It is inherently unstable and insensitive to smaller models or datasets \citep{wei2023skywork}, and further, it lacks correlation with downstream skills to adequately evaluate the model. Addressing these two challenges will help propel further research and practices in the pretraining stage.

In the alignment stage, challenges arise regarding data \citep{albalak2024survey}, long context \citep{niah2023v0} and RLHF effectiveness \citep{wang2024secrets,xu2024dpo}.
SFT gets major parts of things done, but is to some extent sensitive to data quality and composition \citep{liu2024what}. It is tricky to balance the learning of different prompt categories and specific application data.
Various useful applications of LLMs and multi-modal large models require sequence lengths far beyond several thousand \citep{reid2024gemini}. Pretraining on large corpus of long data (context length of tens or hundreds of thousand) turns out inefficient and one desideratum is to extend the context length with minimal continual pretraining and SFT at reasonable costs.
RLHF in open-sourced models has not yet fulfilled the presumed promise as \citet{achiam2023gpt}. Much remains under-explored in terms of RM data, RM training, PPO data and PPO training, etc.

In response to those challenges, we devoted substantial efforts to our LLM models, the 360Zhinao series, and presented the details in this technical report. The 360Zhinao model comprises a base model trained from scratch and a chat model using alignment techniques. In the pretraining stage, we explored the data strategies and their evaluation. First, we built a data cleaning and filtering pipeline. By crawling massive web pages, we employed a series of filtering and cleaning steps. Subsequently, we explored multi-level deduplication and data mixing strategies, ultimately obtaining a corpus of 3.4 TB tokens with high data efficiency. Second, to validate the effectiveness of our data strategies, we constructed a stable and sensitive ablation environment, establishing a set of custom benchmarks associated with downstream skills. Based on this environment, we reported the ablation results of data strategies, effectively guiding the direction of data iterations.

Emphasis is also laid on data in the alignment stage. Initially we prioritized quantity over quality and later the opposite. We now grow SFT data at a more steady pace with higher standards. 
We explored different ways of context window extension and have converged to the simple method of RoPE-base change with tailored data.
We stabilized RLHF training with data and codebase improvement and have applied successfully to specific tasks. The resulting RM has also been used in other data selection and judgement pipelines.

Due to the complexity of LLMs, some crucial intermediate information on model production is absent in existing literature. In this report, we made efforts to break barriers and hope to offer new perspectives to researchers. We have released the implementations of the base- and chat-models to the open-source community, aiming to enhance the openness and transparency of LLMs and facilitate collaboration and reproducibility. Based on these efforts, the contributions of the release of the 360Zhinao models can be summarized as follows:
\begin{itemize}
\item In the pretraining stage, we provide detailed insights into the iteration of data recipes and the construction of an effective ablation experimental environment.
\item In the alignment stage, we present our data-centric approach to data improvement, our converged approach to context window extension and positive findings from RLHF.
\end{itemize}

\section{Pretraining}
Pretraining serves as the foundational stage of developing LLMs, providing them with the capability to learn general linguistic patterns from vast text corpora. This section explores a comprehensive ablation study of data recipes, including cleaning and deduplication methods. We present the model training, as well as evaluation results on benchmarks. 

\subsection{Data}
\begin{figure}[!t]
  \centering
  \includegraphics[width=0.8\textwidth]{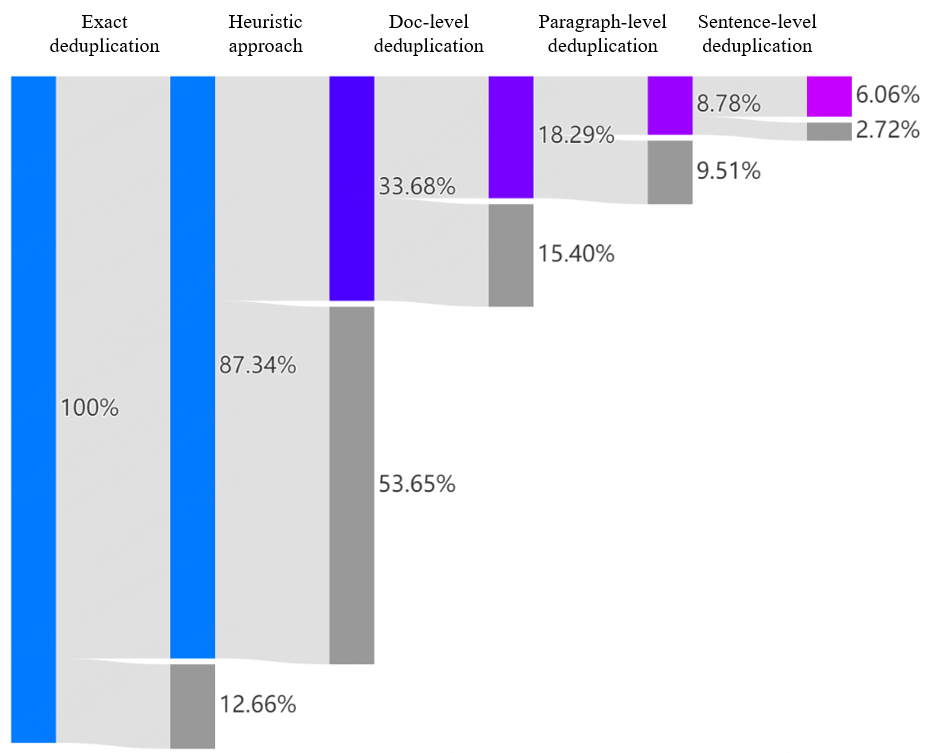}
  \caption{The cascaded retention changes in the web page pipeline.}
  \label{fig:pipeline}
\end{figure}

The publicly accessible web pages serve as one of the primary components of our corpora, including billions of various pages. We have built a data recipe pipeline and performed various meticulous data strategies. The pipeline can be divided into four stages: 1) Data preparation; 2) Data cleaning; 3) Data deduplication and 4) Data mixture. The cascaded changes in data size in the pipeline are shown in Figure \ref{fig:pipeline}. To verify the effectiveness of the data strategy, we constructed a sensitive evaluation environment and conducted a detailed ablation analysis.

\subsubsection{Data preparation}
Faced with such a large scale of web pages, we need to choose the appropriate documents for data cleaning. The pipeline for data preparation includes 1) Url and keyword filtering; 2) Exact deduplication and 3) Language filtering. All actions performed in this pipeline are document-level filtering, removing 12.66\% of the data.

\paragraph{Url and keyword filter}

The uneven quality of websites necessitates that web page filtering must be strategic. We use a combination of black-and-white lists for website filtering. The blacklist mainly records pornographic, violent, as well as cheating, and dead links. The whitelist records high-scoring sites that have been manually reviewed. We prioritize scanning whitelisted sites and discarding pages from blacklisted sites. For the remaining pages, we will calculate the website rating based on meta information such as the website's registration status, authority, and historical performance, and select highly trusted web pages according to the rating. In addition, a sensitive keyword library has also been constructed for further filtering of web pages.

\paragraph{Exact deduplication}
The selected web page may contain potential duplicate pages. Url deduplication and document body deduplication have been adopted into the pipeline. Url deduplication can prevent duplicate crawling and parsing of web pages. Document body deduplication can reduce data redundancy in the dataset. 

\paragraph{Language filter}
Language filtering aims to retain the desired target language web pages. We utilized a FastText model \citep{bojanowski2017enriching, wenzek2019ccnet} for identifying the language of web pages. Specifically, an English web page would be filtered out when containing Chinese characters.

\subsubsection{Data cleaning}
The goal of data cleaning is to extract the main content of web pages and obtain high-quality text. The cleaning pipeline integrates heuristic rules and models, including junk text cleaning, content quality filtering, and personally identifiable information (PII) cleaning. Through those three stages of cleaning, we have significantly improved content quality and security, removing a total of 53.65\% of the text.

\paragraph{Junk text filter}

Generally, the content extracted from structured web pages still contains a large amount of junk text, such as web page navigation, copyright notices, related content recommendations, and advertisements. Junk text is all over the place, following similar templates, and lacking diversity, which can hinder the effectiveness and efficiency of model training if not deleted cleanly. We adopted a pipeline for content cleaning that includes paragraph granularity and document granularity, using heuristic rules and classification models. We pursue the accuracy of deleting the junk text without compromising the coherence of the content during the cleaning pipeline.

Paragraph granularity cleaning aims to preserve the main content of web pages while removing irrelevant junk text. We trained a FastText model to identify whether a given paragraph is junk. This model employs a semi-supervised learning approach based on incremental learning, with training data reaching millions in scale after multiple iterations. The F1-score of the junk text model reached 99\%. Given a document, all its paragraphs undergo prediction by this model and then whether a paragraph is discarded depends on the status sequence of all paragraphs.

Document granularity cleaning is performed after paragraph subsequently cleaning to remove documents lacking information or containing formatting issues. Heuristic rules based on frequency and ratio, inspired by Dolma's practice, serve as the primary cleaning methods \citep{soldaini2024dolma, chen2023data, albalak2024survey}. Specific rules can be found in Appendix \ref{sec:app:clean}.

After paragraph and document granularity cleaning, we retained the main content of the web page and removed low-information documents. This cleaning process is trivial but important, it cuts down abundant duplicate data.

\paragraph{Quality filter}
High-quality data can improve the performance of models. However, to our knowledge, there is no unified definition of data quality. Existing literature typically uses classification recognition or perplexity (PPL) truncation to partition high-quality data \citep{brown2020language, young2024yi}. For classification methods, people use Wikipedia and books as high-quality data and use classifiers to identify target documents that are similar to these high-quality data, which may unexpectedly bias the data distribution and damage diversity. PPL uses a language model to calculate the PPL score of a document and then filters it using a specific threshold. However, PPL truncation may risk compromising high-quality data in specific niche domains, such as exams and ancient literature.

In our practice, we refer to the principle that “data that is known to be written by humans and has likely gone through an editing process” \citep{albalak2024survey, gao2020pile} when selecting high-quality data. High-quality data has rich informativeness and good coherence, without any pornographic or violent elements. Based on these criteria, we manually annotated 60k samples using active learning, with a balanced number of positive and negative samples. We trained a classification model based on BERT \citep{devlin2018bert} to rate content quality, with an F1-score of 85\%. According to the quality ratings, low-quality documents were removed during the quality filtering stage.

\paragraph{Personal identifiable information filter}
Personal identity information (PII) of users may be leaked from web pages \citep{kumar2024ethics, subramani2023detecting}. PII will be abundant when the data scale is very large \citep{elazar2023s}. PII recognition can be achieved through model-based methods with better performance or rule-based methods with high computational efficiency \citep{soldaini2024dolma}. In our practice, we used rule-based methods through regular expressions, including email detection, IP address detection, and phone number recognition. The text block that hits the rule will be replaced with a specific token. Although the cleaning of PII does not significantly improve the performance of the model, it enhances the security protection for users in the network.

\subsubsection{Data deduplication}
Within the data recipe pipeline, deduplication is critically important due to the pervasive presence of duplicate content, which arises from various origins, including identical copies, cross-site plagiarism, and reference \citep{penedo2023refinedweb}. Although duplicate data can enhance the model's memorization, it weakens the model's generalization and impedes training efficiency \citep{hernandez2022scaling, lee2023beyond, marion2023less, tirumala2024d4, sachdeva2024train}. Consequently, we devoted considerable efforts to exploring data deduplication. Our strategy is multi-level deduplication, including document-level, paragraph-level, and sentence-level, and detailed ablation experiments were conducted to validate the effectiveness.

\paragraph{Document deduplication}
In document deduplication, we employed the Minhash \citep{wu2020review} combined with the Locality Sensitive Hashing (LSH) \citep{jafari2021survey} method, which maps each document to a unique hash value with fixed length. This approach captures local features of the text, effectively identifying duplicate documents. In our implementation, the hash value was set to 128 bits. Document deduplication removed 15.40\% of duplicate documents. Ablation experiments demonstrate a significant improvement in model performance due to document deduplication, with detailed data available in Appendix \ref{sec:app:doc_dedup}.

\paragraph{Paragraph deduplication}
Paragraph deduplication marks identical paragraphs across pages as duplicates, employing an exact match. We observed that paragraph deduplication removes a significant number of high-frequency template-like spans, such as homogeneous advertisements and recommendations, which to some extent compensate for the oversight of the junk text model. Paragraph deduplication can improve the corpus diversity but may compromise the text coherence, which is an accompanying side effect. To assess the impact of diversity and coherence on the model, we conducted a series of ablation experiments. Specifically, we aggregated each set of duplicate paragraphs into a group and deleted 10\%, 20\%, $\cdots$, and 50\% of elements within the group to control the deduplication intensity. Greater intensity implies better data diversity. We found that as the deletion ratio increased from 10\% to 30\%, the model's performance steadily improved; however, further increasing the deletion ratio to 50\% did not enhance the model's performance and even led to a slight decline. This suggests that: 1) the impact of diversity is more significant than the impact on coherence; 2) The growth in diversity tends to saturate at a deletion ratio of 30\%. Further increases do not promote model performance and may cut down data quantity. Therefore, we ultimately chose a 30\% deduplication ratio for duplicate paragraphs to optimize the dataset. Detailed ablation results can be found in Appendix \ref{sec:app:para_dedup}.

\paragraph{Sentence deduplication}

Recognizing the potential of paragraph deduplication in enhancing the training of LLMs, we have further refined our approach by implementing sentence deduplication. This strategy targets the reduction of redundant text fragments found across web pages, a known factor that can impair model training, as highlighted in literature \citep{penedo2023refinedweb, hernandez2022scaling}. To address this, we have adopted a more stringent sentence deduplication strategy. Specifically, we count the frequency $N$ of each sentence and retain only the square root of $N$ unique sentences through deduplication. Moreover, to maintain coherence, we employ a sliding n-gram (where n=16) concatenation technique, thus mitigating the risk of damaging excessively short sentences. Our ablation experiments unequivocally demonstrate the positive impact of sentence deduplication on model performance. detailed results and analysis can refer to Appendix \ref{sec:app:sent_dedup}.

\subsubsection{Data mixture}
Beyond web data, our corpus encompasses some high-quality sources, including Wikipedia, books, code stacks, and various domain-specific documents spanning legal, financial, and ancient texts, etc. Mixing these multi-source data for pretraining corpora presents a challenge \citep{ye2024data, chen2024skill, shen2023slimpajama}. In this report, we considered the following three aspects. First, data blending needs to balance diversity and quality. We found that compared to high-quality knowledge-intensive resources like Wikipedia, web data exhibit better diversity and quantity but inferior quality. Excessive skewness towards web data may degrade model performance. In such cases, mitigating data skewness can be achieved by appropriately oversampling knowledge-intensive data. Generally, oversampling ratios typically do not exceed 5 according to literature \citep{ye2024data}. Second, data blending needs to balance data efficiency and quantity. Data efficiency refers to the improvement of the model performance when trained on the same scale of data. Higher proportions of high-quality data can enhance data efficiency, yet such data is finite and thus necessitates down-sampling of lower-efficiency data. Nevertheless, empirical evidence from scaling laws underscores the critical role of data quantity in determining a model’s ultimate performance. Additionally, data blending needs to consider the granularity of the mixture, with options ranging from source-level, topic-level, to content-type granularity. Finer-grained mixing can finely enhance the diversity of the blended data.

In our practice, on one hand, we applied meticulous cleaning strategies and multi-level deduplication to web data to compress data scale and enhance its efficiency. On the other hand, we balanced web data by moderately oversampling knowledge data. Furthermore, we have experimented with data mixing approaches by employing Doremi-based sampling techniques \citep{xie2024doremi}, as well as source- and topic-level mixing strategies geared towards maximizing diversity. Unfortunately, up until now, these explorations have not yielded statistically significant improvements. Our open-source model is based on a heuristic grid search to determine the ratios of each source, validated through experiments to utilize the optimal ratios. The distribution of data ratios is depicted in Table \ref{table:mixture}. In the future, we envision further investigation into data mixture, which may harbor substantial improvement potential.

\begin{table}
\centering
\centering
    \small
    \renewcommand{\arraystretch}{1.25}
    \caption{The data mixture for pretraining.}
\begin{tabu}{c|c} 
\hline
Source           & Percentage  \\ 
\hline
Webpages         & 63.33\%     \\ 
\hline
Code             & 9.25\%      \\ 
\hline
Math             & 7.31\%      \\ 
\hline
Books            & 6.20\%      \\ 
\hline
Patents          & 2.85\%      \\ 
\hline
Academic Papers~ & 2.38\%      \\ 
\hline
Encyclopedia     & 0.96\%      \\ 
\hline
Other            & 7.72\%      \\
\hline
\end{tabu}
\label{table:mixture}
\end{table}

\subsection{Tokenizer}
We focus on both the text compression rate of the tokenizer and its performance in downstream tasks. High compression rates can improve both training and inference efficiency. Considering encoding efficiency, we follow GPT-3.5 and GPT-4 \citep{brown2020language, achiam2023gpt}, employing the efficient Byte Pair Encoding (BPE) tokenizer, tiktoken \citep{tiktoken}, for tokenization \citep{bai2023qwen}. Given that OpenAI's cl100k vocabulary has shown good compression performance in English and many other languages, we adopt cl100k as our base vocabulary. However, this base vocabulary does not perform well in Chinese. Therefore, we augment the vocabulary with commonly used characters and words from Chinese and some other languages. Following Llama \citep{touvron2023llama}, we split numbers into individual digits to address inconsistent number splitting. Considering the prevalence of whitespace indentation in code data, we also define specialized tokens representing varying levels of space indentations, namely [space2], [space3], [space4], and [space8]. Consequently, our final vocabulary size amounts to 158k entries.

\begin{figure}[!t]
  \centering
  \includegraphics[width=1.0\textwidth]{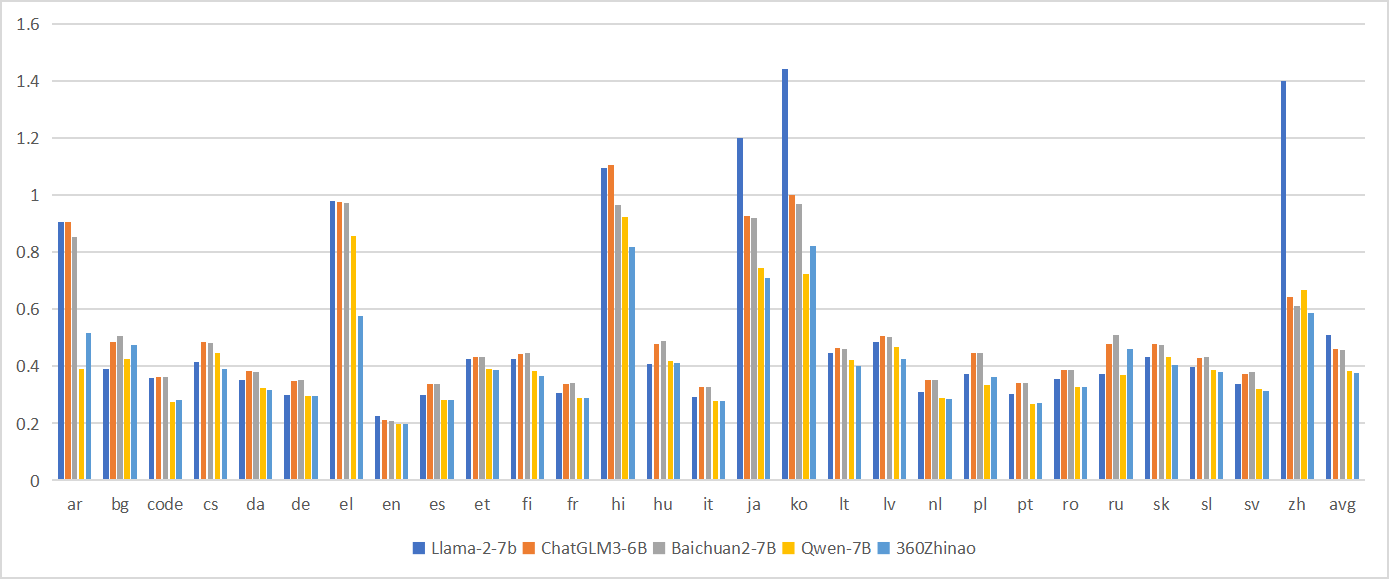}
  \caption{The comparison of compression rates among tokenizers.}
  \label{fig:tokeizer}
\end{figure}

Our tokenizer's performance in compression rate is illustrated in Figure \ref{fig:tokeizer}, where it is compared with tokenizers of Llama-2-7b \citep{touvron2023llama2}, ChatGLM3-6B \citep{zeng2022glm}, Baichuan2-7b \citep{yang2023baichuan}, and Qwen-7B \citep{bai2023qwen}. The results indicate that our tokenizer exhibits superior performance in compression rate.

\subsection{Model architecture}
Similar to most open-source models, the 360Zhinao model is based on the transformer architecture and roughly follows the LLama structure \citep{touvron2023llama2}. It adopts the Pre-Norm structure with RMSNorm \citep{zhang2019root} function and utilizes SwiGLU \citep{shazeer2020glu} as the activation function for the feed-forward network (FFN). The intermediate layer dimension is set to 8/3 times the hidden size and rounded up to 256. Additionally, Rotary Embedding \citep{su2024roformer} is employed for positional encoding. Notably, we choose to use FP32 precision for the inverse frequency matrix instead of half-precision to achieve higher accuracy. For the entire model, we remove biases according to \citep{chowdhery2023palm}, except for the biases in the attention QKV layers, to enhance the model's extrapolation capability \citep{su2024roformer}.

\subsection{Training detail}
We adhere to the standard autoregressive language modeling approach \citep{radford2018improving} for training the 360Zhinao model, which involves predicting the next token based on the preceding ones. For data organization, we concatenate documents within the same source, truncate them to a specified length, and shuffle them thoroughly, all within the training framework. We train models with a context length of 4096 and employ Flash Attention \citep{dao2023flashattention} in the attention module, effectively enhancing computational efficiency and reducing memory usage. 

For optimization, we utilize AdamW \citep{loshchilov2017decoupled} for pre-training, with hyperparameters set as follows: $\beta_1=0.9$, $\beta_2 = 0.95$, $\epsilon = 1E-8$, and weight decay $= 0.1$. Notably, the hyperparameter $\epsilon$ in AdamW is set to $1E-5$ in Llama2 \citep{touvron2023llama2},  which reveals gradient instability issues through our experiments. Experimental analysis in \citep{wortsman2023small} explores the impact of this parameter on training and suggests appropriate reduction when scaling up the model size. Additionally, we do not employ dropout \citep{chowdhery2023palm} during the pre-training stage. We adopt a cosine learning rate schedule with warmup, where the learning rate decays to 10\% of the maximum learning rate by the final training step \citep{scao2022language}. To ensure training stability, we have experimented with various tricks on FP16 training, including embedding norm \citep{scao2022language}, QK layernorm \citep{dehghani2023scaling}, normhead \citep{yang2023baichuan}, etc. These methods essentially constrain potentially unstable forward activations using normalization operations. However, with the adoption of BF16, training models at the billion-parameter scale appear to exhibit less instability. Therefore, we adopt BF16 mixed precision as the standard for training, despite a slight sacrifice in memory usage. Grad clipping is set to 1.

\subsection{Benchmark evaluation}
We trained a 360Zhinao-7B-base model with 7 billion parameters on a corpus of 3.4 trillion tokens. In this section, we report the performance of our base model on standard benchmarks against other similarly scaled open-source models, such as Baichuan2-7B \citep{yang2023baichuan} and DeepSeek-7B \citep{bi2024deepseek}. We evaluate the models across various dimensions, including natural language understanding, knowledge, mathematics, code generation, and logical reasoning, using eleven popular benchmarks including C-Eval \citep{huang2024c} and AGIEval \citep{zhong2023agieval}, etc. Evaluation experiments are conducted in zero-shot or few-shot settings. Specifically, we adopt a 5-shot approach for C-Eval, MMLU \citep{hendrycks2020measuring}, and CMMLU \citep{li2023cmmlu}, a 4-shot approach for MATH \citep{hendrycks2021measuring} and GSM8K \citep{cobbe2021training}, a 3-shot approach for BBH \citep{suzgun2022challenging}, a 1-shot approach for MBPP \citep{austin2021program}, and 0-shot for others. For fairness in comparison, we strictly adhere to the official implementation of OpenCompass \citep{contributors2023opencompass}, and gathered baseline scores from the official OpenCompass platform.

\begin{table}[!t]
    \centering
    \small
    \setlength\tabcolsep{2pt} 
    \renewcommand{\arraystretch}{1.25}
    \caption{Comparison of results on popular benchmarks.}
    \label{table:benchmark}
    \begin{tabular}{c|c|c|c|c|c|c|c|c|c|c}
    \hline
        Model & Avg & CEval & AGIEval & MMLU & CMMLU & HellaSwag & MATH & GSM8K & HumanEval & MBPP \\ \hline
        Baichuan2-7B & 41.49 & 56.3 & 34.6 & 54.7 & 57 & 67 & 5.4 & 24.6 & 17.7 & 24 \\ \hline
        Baichuan-7B & 31.94 & 44.7 & 24.6 & 41.5 & 44.6 & 68.4 & 2.5 & 9.6 & 9.1 & 6.4 \\ \hline
        ChatGLM3-6B & 58.67 & 67 & 47.4 & 62.8 & 66.5 & 76.5 & 19.2 & 61 & 44.5 & 57.2 \\ \hline
        DeepSeek-7B & 39.8 & 45 & 24 & 49.3 & 46.8 & 73.4 & 4.2 & 18.3 & 25 & 36.4 \\ \hline
        InternLM2-7B & 58.01 & 65.7 & 50.2 & 65.5 & 66.2 & 79.6 & 19.9 & 70.6 & 41.5 & 42.4 \\ \hline
        InternLM-7B & 39.33 & 53.4 & 36.9 & 51 & 51.8 & 70.6 & 6.3 & 31.2 & 13.4 & 14 \\ \hline
        LLaMA-2-7B & 33.27 & 32.5 & 21.8 & 46.8 & 31.8 & 74 & 3.3 & 16.7 & 12.8 & 14.8 \\ \hline
        LLaMA-7B & 30.35 & 27.3 & 20.6 & 35.6 & 26.8 & 74.3 & 2.9 & 10 & 12.8 & 16.8 \\ \hline
        Mistral-7B-v0.1 & 47.67 & 47.4 & 32.8 & 64.1 & 44.7 & 78.9 & 11.3 & 47.5 & 27.4 & 38.6 \\ \hline
        MPT-7B & 30.06 & 23.5 & 21.3 & 27.5 & 25.9 & 75 & 2.9 & 9.1 & 17.1 & 22.8 \\ \hline
        Qwen1.5-7B & 55.12 & 73.57 & 50.8 & 62.15 & 71.84 & 72.62 & 20.36 & 54.36 & 53.05 & 36.8 \\ \hline
        Qwen-7B & 49.53 & 63.4 & 45.3 & 59.7 & 62.5 & 75 & 13.3 & 54.1 & 27.4 & 31.4 \\ \hline
        XVERSE-7B & 34.27 & 61.1 & 39 & 58.4 & 60.8 & 73.7 & 2.2 & 11.7 & 4.9 & 10.2 \\ \hline
        Yi-6B & 47.8 & 73 & 44.3 & 64 & 73.5 & 73.1 & 6.3 & 39.9 & 15.2 & 23.6 \\ \hline
        360Zhinao-7B & 56.15 & 74.11 & 49.49 & 67.44 & 72.38 & 83.05 & 16.38 & 53.83 & 35.98 & 42.4 \\ \hline
    \end{tabular}
\end{table}

The comparative results are presented in Table \ref{table:benchmark}. Compared with models of similar size, the 360Zhinao-7B-base model demonstrates outstanding overall performance. It achieves top rankings in multiple benchmark assessments, such as C-Eval, MMLU, HellaSwag, and LAMBADA, showcasing strong competitiveness in both Chinese and English knowledge and reasoning comprehension abilities. These results highlight the model's superior capacity to handle diverse linguistic and cognitive tasks within the realms of natural language processing and understanding.

\subsection{Ablation study}
As is well known, data determines the upper limit of model performance. Therefore, we have implemented meticulous strategies in data recipes. The effectiveness of these strategies will be validated through ablation experiments. 

\subsubsection{Ablation design}
Ablation evaluation aims to validate the effectiveness of strategies in small-data experimental scenarios. We establish a tripartite evaluation framework. First, a validation set is created to monitor its loss during the training process. Second, beyond the OpenCompass evaluation, we develop a more stable custom benchmark. Finally, supervised fine-tuning (SFT) evaluation is introduced, which can directly correlate with downstream task performance with a tangible measure.

Typically, the validation set is sampled from the training dataset. To ensure the diversity of the validation set, we employ sampling at the topic granularity and ensure that there is no overlap between the validation and the training data. The validation set is utilized to assist in assessing model performance by valid loss and make early stopping decisions during the training stage.

Ablation experiments are conducted on small models and limited tokens to sensitively evaluate the effectiveness of data strategies. Generally, OpenCompass is a widely used benchmark to assess zero/few-shot generation capabilities. However, we encountered two issues: 1) instability and insensitivity of OpenCompass to subtle model differences on small-scale ablation, and 2) difficulty of aligning OpenCompass with downstream-specific business capabilities. To address these challenges, we made two efforts. First, we systematically curated a subset from OpenCompass, retaining effective and sensitive evaluations. Second, we developed a custom benchmark, named 360Eval, which is more suitable for ablation and aligns with downstream performance. 360Eval contains a hierarchical skill list, including machine reading comprehension, classification, extraction, translation, examination, and summarization. For each skill, we collected multiple publicly available datasets to ensure the diversity of the evaluation. To make metrics more robust and effective, we enhance the diversity of prompts by constructing multiple templates. Additionally, we improve the way of metric computation by adopting a generative approach. For example, in multiple-choice questions, the generation of answer options (e.g., A, B, C, D) is supplemented by generating answer content \citep{zheng2023large}. Finally, we reduce the sample size at the task granularity to balance evaluation efficiency and error. Regarding potential contamination, GPT-3 indicates that low levels of contamination do not lead to unfairness issues without specific intervention. Our experimental findings concur with this conclusion. Therefore, the decontamination has been omitted to help alleviate computational costs during data iterations.

As a pivotal component of our evaluation, the SFT evaluation, despite its high computational cost, directly reflects the downstream effects of the base model. First, we fine-tune the base model to a chat model and then utilize a trained reward model to score the generated results by the chat model. Finally, we conduct evaluation comparisons of the ablation experiments from three dimensions: 360Eval, OpenCompass, and SFT evaluation.

\subsubsection{Ablation setting}
The ablation experiments were conducted on a 1.8B model and trained on 100B tokens. Carefully constructed training data is essential for ablation experiments. We set a fixed, diverse mixed dataset as a base training set. The base data is fixed to facilitate comparisons across multiple strategies, while its diversity ensures the model's basic effectiveness. To enhance the discriminability of the ablation strategies, during the ablation experiments, our training data consists of 50\% experimental data and 50\% base data. A higher proportion of experimental data enhances the discriminability of the ablation strategies. 

Based on this experimental setup, we conducted ablation analyses on document, paragraph, and sentence deduplication. Detailed experimental results can be found in Appendix \ref{sec:ablation}.

\section{Alignment}
Following common practice \citep{ouyang2022training,bai2022training,touvron2023llama,bai2023qwen,yang2023baichuan,bi2024deepseek}, we performed SFT and RLHF on the pretrained models.

For SFT, we continuously improved our training data. We initially had over 10 million prompt-answer pairs and iteratively reduced the number to hundreds of thousand while improving their quality. We also constructed PoT \citep{chen2023program} data for math and reasoning tasks.
We specifically extended the context length from 4K to 32K and 360K during finetuning with simple RoPE-base changes and constructed long SFT data \citep{xiong2023effective,liu2024world}.

For RLHF, we continuously worked on its infrastructure to support PPO on larger models with minimal hardware. We achieved a little improvement on general LLM tasks and more significant gains in specific areas such as translation and agentic tasks.

Human annotators play an important role throughout the alignment stage. They provide high-quality SFT data, preference data and objective comparison of different model versions. We have built an annotation team of around 20 people, and worked closely with them in the loop on each annotation task to support 360's AI products such as 360 AI-search\footnote{\url{https://so.360.com/}} and 360 AI-browser\footnote{\url{https://browser.360.cn/ai/}}.

\subsection{SFT}

We took a data-centric approach to SFT and continuously improved the SFT dataset, firstly in quantity and then in quality. PoT data is specifically constructed for math and reasoning tasks to mitigate the dissatisfaction of 7B-sized models to compute and reason.
Those efforts resulted in the competitive performance of 360Zhinao-7B-Chat-4K on relevant benchmarks.

Building upon the 4K model, we extended its context length to 32K and 360K with simple RoPE-base changes and carefully curated long data. The 32K and 360K models exhibit top-tier performance on LongBench \citep{bai2023longbench}
and near-perfect all-green results on NIAH (Needle-In-A-Haystack evaluation) \citep{niah2023v0,niah2023v1}.

\subsubsection{SFT data}
\label{sec:sft.data}

\textbf{Quantity-Quality:} 
Early in 2023 when we just started training 360Zhinao models, we prioritized quantity over quality of the SFT data. Back then, we expanded our SFT dataset every time one batch of human annotation is ready or one open-sourced dataset is accessible, and model performance did improve with such expansion.

However, such procedure quickly became impractical in mid 2023. The sheer number of over 10 million prompt-answer pairs prohibited iterative data cleaning and quality improvement. Emphasis was therefore shifted from quantity to quality, aiming to limit total SFT data size within hundreds of thousand to facilitate continuous efforts.

Open-sourced English data are generally of higher quality than Chinese counterparts. We empirically found that adding some of high-quality open-sourced English SFT data \citep{mukherjee2023orca,OpenOrca,mitra2024orcamath,dbrx2023data} has positive effects even on our internal pure-Chinese evaluation set.
We posit this is because pretrained LLMs are inherently aligned across languages exhibiting translation capabilities, and training on instruction-following data in one language benefits alignment in all languages.
We generally have less than 5\% English SFT data in the training dataset.

\textbf{Quality processing:}
To prune the SFT data, we first grouped them by their original sources and rated each source manually on a randomly sampled subset. Data from sources rated below a certain threshold were directly discarded.

We then leveraged the 360Zhinao model itself to label the categories and difficulty levels of the remaining data. Such labelling is achieved by first asking our annotation team to label a seed set in category and difficulty level and then fine tuning a 360Zhinao model specifically on this seed set. The model could then label all the data as initial annotation.

\textbf{Current status:}
We have been continuously working on the resulting data to correct the category and difficulty labels, prune/reweight the data according to the labels and improve the quality of the prompts and answers therein.
Unreasonable prompts were removed or modified and answers were refined and reformatted.
This is an endless process and we always find some of us on it. Upon release of 360Zhinao-7B v1.0, we have around 600 thousand SFT data with manageable size of each category for us to iterate over and improve upon.
New high-quality data are still added to this dataset on a regular basis, but we now keep dataset expansion at a more steady pace with higher standards on quality.

\textbf{PoT:}
For math and reasoning tasks, we let our model write Python code to compute instead of generating the computational output directly with natural language, similar to \citet{chen2023program}.
Such SFT data were constructed with relevant prompts and manually verified one-by-one by specifically delegated colleagues to ensure correctness.
Special tokens were added right before and after the Python code in the SFT data so that the generated code could be identified by a separate code interpreter in deployment to get the executed results.

\subsubsection{4K Model Training and Evaluation}
The released 360Zhinao-7B v1.0 model (360Zhinao-7B-Chat-4K) was trained with standard SFT, with data packed into the 4K window. We have tried separating attention masks of different data with $\texttt{flash\_attn\_varlen\_func}$ but found it inferior to naive packing, possibly because naive packing approximately induces the desired multi-turn conversation capabilities.
Besides naive packing, we also constructed around 5,000 multi-turn conversations with role exchange prompting \citep{ding2023enhancing} and role-playing ghost attention \citep{touvron2023llama2} to explicitly induce multi-turn capabilities. Some other data in our dataset are also naturally multi-turn.
  

Table \ref{tab:mtbench} demonstrates competitive results of our released 4K model on MT-bench \citep{zheng2024judging}, a benchmark with 80 high-quality multi-turn dialogue questions to test the capabilities of multi-turn dialogue and instruction following. MT-bench consists of 8 common user prompt categories: writing, role-play, extraction, reasoning, mathematics, coding, knowledge I (STEM), and knowledge II (humanities/social sciences). For each category, 10 multi-turn questions were manually designed, each turn has 2 questions, and the answers are scored using GPT-4 (accessed on Apr. 8, 2024).

\begin{table}
    \centering
    \begin{tabular}{l|c|c|c} \hline 
        \textbf{Model} & \textbf{Turn1} & \textbf{Turn2} & \textbf{average}\\ \hline 
        Qwen-7B-Chat & \textbf{6.5725} & 5.4000 & 5.9862\\ 
        Baichuan2-7B-Chat & 6.4562 & 5.5562 & 6.0062\\ 
        InternLM-7B-Chat & 5.5625 & 4.0696 & 4.8207\\ 
        360Zhinao-7B-Chat & 6.5062 & \textbf{5.8762} & \textbf{6.1962}\\ \hline
    \end{tabular}
    \caption{MTBench evaluation results.}
    \label{tab:mtbench}
\end{table}

\begin{figure}
\begin{center}
    \includegraphics[width=0.95\linewidth]{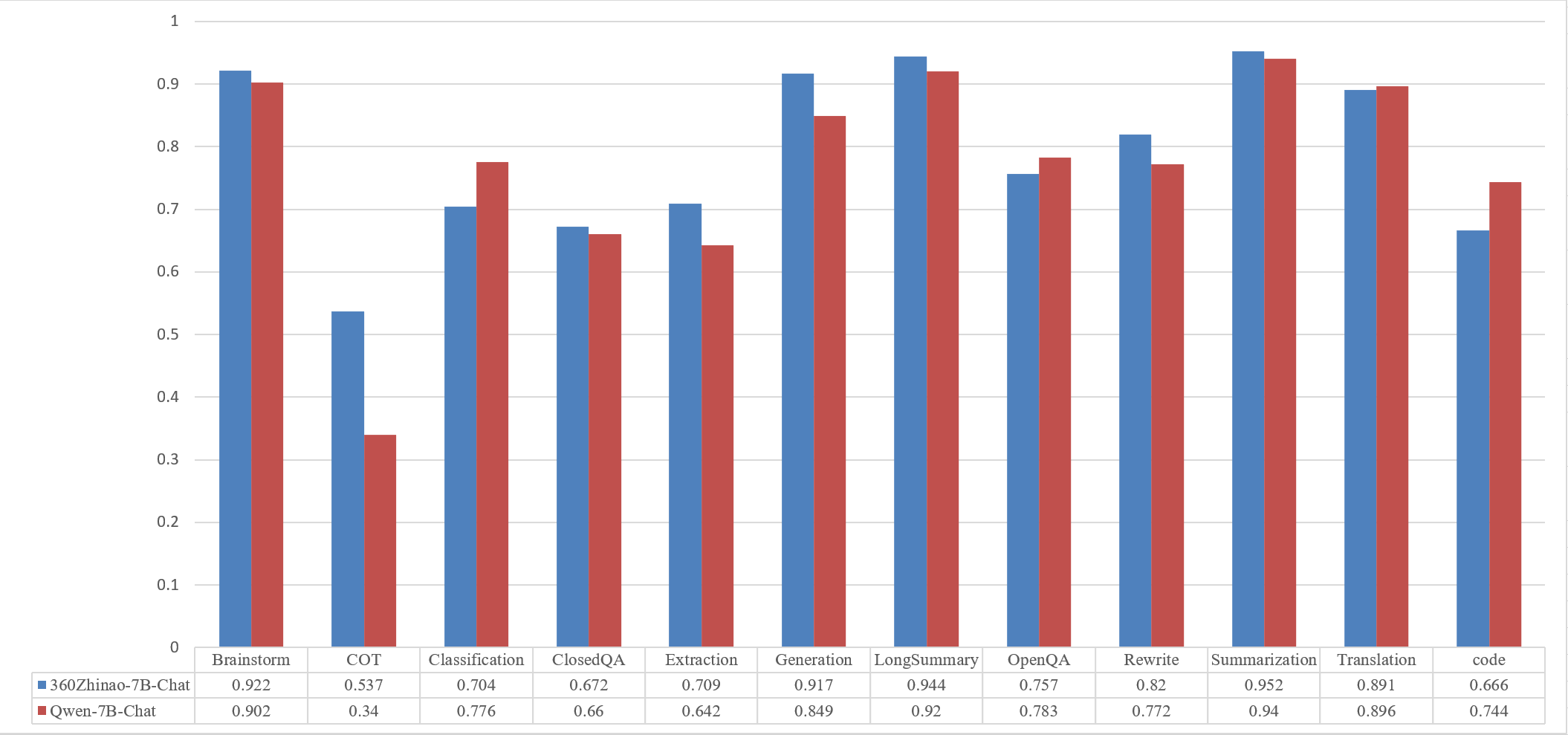}
\end{center}
\caption{Results on the internal evaluation set of 360Zhinao-7B-Chat and Qwen-7B-Chat. 360Zhinao-7B-Chat outperforms Qwen-7B-Chat on most prompt categories.}
\label{tab:sft.internal}
\end{figure}

We also compared against Qwen-7B-Chat (Qwen v1.0) on our internal evaluation set consisting of 1,100 prompts covering tasks of OpenQA, ClosedQA, summarization, COT questions and code, etc (similar to the prompt categories in \citet{ouyang2022training}).
360Zhinao-7B-Chat achieved an average score of 0.79 compared to Qwen-7B-Chat's 0.77. Detailed scores are shown in Figure \ref{tab:sft.internal}.

\subsubsection{32K and 360K Models}

We extended the context window of 360Zhinao-7B (base model) to 32K and 360K in the SFT stage with minimal costs.
To train the long-context models, we adopted a two-stage approach:

\begin{itemize}
\item \textbf{First stage:} We increased RoPE base and extended the context length to 32K.

Firstly, we performed Continual Pretraining on approximately 5B tokens with a 32K context window.

Then we finetuned the model using long data from various sources, including high-quality human-labeled 32K data, together with the general SFT data in Sec. \ref{sec:sft.data}.

\item \textbf{Second stage:} We extended the context length to 360K, training with the following data:

1. A small number of high-quality human-labeled super-long data.

2. Due to the scarcity of annotated super-long data, we constructed various forms of synthetic data:

\begin{itemize}
\item Multi-Doc QA: Similar to \citet{junqing2023never}, we generated multi-document QA pairs based on our own database. Multiple QA pairs are constructed for one row of Multi-Doc QA data input, resulting in a multi-turn format and significantly improving the training efficiency.
\item Single-Doc QA: Similar to \citet{xiong2023effective}, we constructed multi-turn QA data based on different segments within one row of long-text input.
\end{itemize}

We initially trained 360K models with Megatron-LM \citep{shoeybi2019megatron,korthikanti2023reducing} after optimizing its GPU memory consumption by removing unnecessary attention masks and changing to in-place operations.
However, Megatron's tensor parallelism scales poorly with sequence length, and its context parallelism wasn't quite ready back when we were choosing from training options. We have now switched to Ring Attention \citep{li2021sequence,liu2023ring}, which requires much fewer machines and has similar overall training efficiency.

\end{itemize}

\textbf{Method:} 
Initially, we increased RoPE base from 10,000 to 1,000,000 in the first stage, and then from 1,000,000 to 50,000,000 in the second stage. This gives satisfactory results on LongBench \citep{bai2023longbench} with the 32K model and on NIAH with the 360K model, accompanied by however notable degradation with the 360K model on LongBench.
Therefore, we chose to use RoPE base of 50,000,000 in both 32K and 360K models. We empirically found that this had no impact on 32K performance, but significantly mitigated degradation of the 360K model on 32K tasks.

\textbf{LongBench:} 
We report evaluation results on LongBench \citep{bai2023longbench} in Table \ref{tab:longbench}. 360Zhinao-7B-Chat-32K achieved the best score among models within 10B. By using RoPE base of 50,000,000 directly in the first stage, 360Zhinao-7B-Chat-360K mostly retained its 32K performance and only degraded significantly on code, mainly because our second-stage data to train the open-sourced 360Zhinao-7B-Chat-360K model had no code. We could improve its code performance simply by adding code data in the second stage in later experiments.

\begin{table}
    \centering

\begin{tabular}{m{0.2\linewidth} |m{0.1\linewidth} |m{0.1\linewidth} |m{0.1\linewidth} |m{0.1\linewidth} |m{0.1\linewidth} |m{0.1\linewidth} } \hline 
\textbf{Model} & \textbf{Avg} & \textbf{Single-Doc QA} & \textbf{Multi-Doc QA} & \textbf{Summa-rization} & \textbf{Few-shot Learning} & \textbf{Code Completion} \\ \hline
GPT-3.5-Turbo-16k & 37.84 & 61.2 & 28.7 & 16 & 29.2 & 54.1 \\ \hline 
ChatGLM2-6B-32k & 37.16 & 51.6 & 37.6 & 16.2 & 27.7 & 52.7 \\ \hline 
ChatGLM3-6B-32k & 44.62 & \textbf{62.3} & 44.8 & 17.8 & 42 & 56.2 \\ \hline 
InternLM2-Chat-7B & 42.20 & 56.65 & 29.15 & \textbf{17.99} & 43.5 & \textbf{63.72} \\ \hline 
Qwen1.5-Chat-7B & 36.75 & 52.85 & 30.08 & 14.28 & 32 & 54.55 \\ \hline 
Qwen1.5-Chat-14B & 39.80 & 60.39 & 27.99 & 14.77 & 37 & 58.87 \\ \hline 
360Zhinao-7B-Chat-32K & \textbf{45.18} & 57.18 & \textbf{48.06} & 15.03 & \textbf{44} & 61.64 \\ \hline 
360Zhinao-7B-Chat-360K & 39.25 & 52.82 & 48.01 & 14.4 & 41.25 & 39.75 \\ \hline
\end{tabular}
    \caption{LongBench evaluation results. 360Zhinao-7B-Chat-32K achieved the best score among models within 10B. The 360K model only degraded significantly on code, mainly because the 360k-stage data of this version had no code. Later experiments with code data helped resolve this.}
    \label{tab:longbench}
\end{table}

\textbf{NIAH:}
Besides retaining 32K performance in the 360K model, we mainly focused on improving NIAH (Needle-In-A-Haystack evaluation) \citep{niah2023v0,niah2023v1} within the 360K-token context window.
Although NIAH is a somewhat artificial and unnatural task, it still serves as a necessary check for long-context models. Unsatisfactory NIAH results cast shadow on real-world long context applications.

It's worth mentioning that in our experiments, we found the value retrieval variant \citep{niah2023v1} of NIAH (widely used recently in \citet{reid2024gemini,liu2024world}) is relatively easy and we could quickly get 100\% perfect all-green results. However, the model checkpoints with all-green results on the value retrieval variant still struggles with the original version \citep{niah2023v0} of NIAH.
We posit that the needle format of the value retrieval variant, "The special magic \{random city\} number is: \{random 7-digit number\}.", is much more distinct than the original-version needle "The best thing to do in San Francisco is ..." in the haystack of natural text.
The commonly used haystack of Paul Graham essays discusses Silicon Valley and inevitably San Francisco, which poses even more confusion with the original-version needle.

\begin{figure}
    \centering
    \parbox{0.3\columnwidth}{\centering
    \includegraphics[width=0.3\columnwidth]{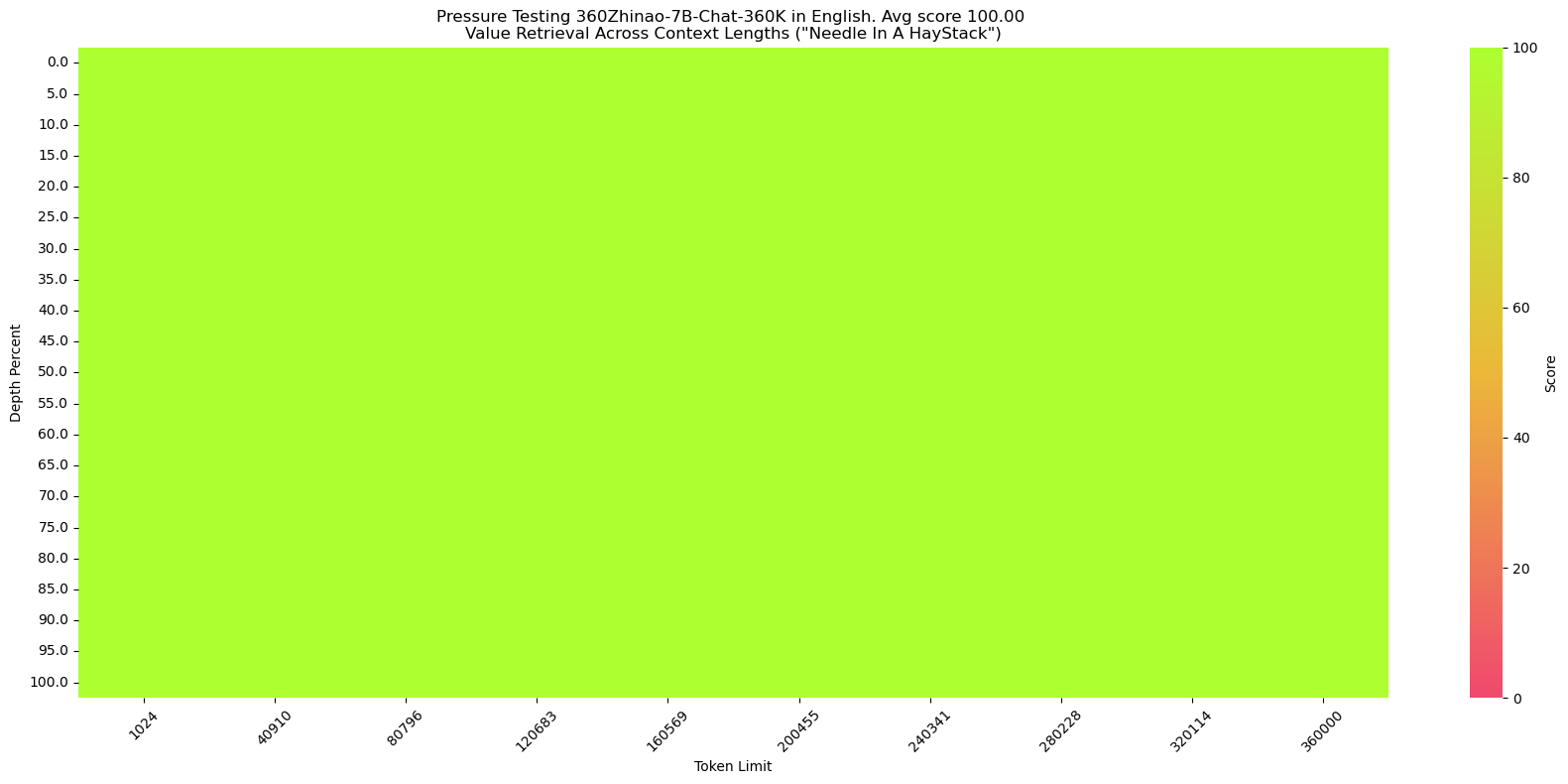}
    }
    \parbox{0.3\columnwidth}{\centering
    \includegraphics[width=0.3\columnwidth]{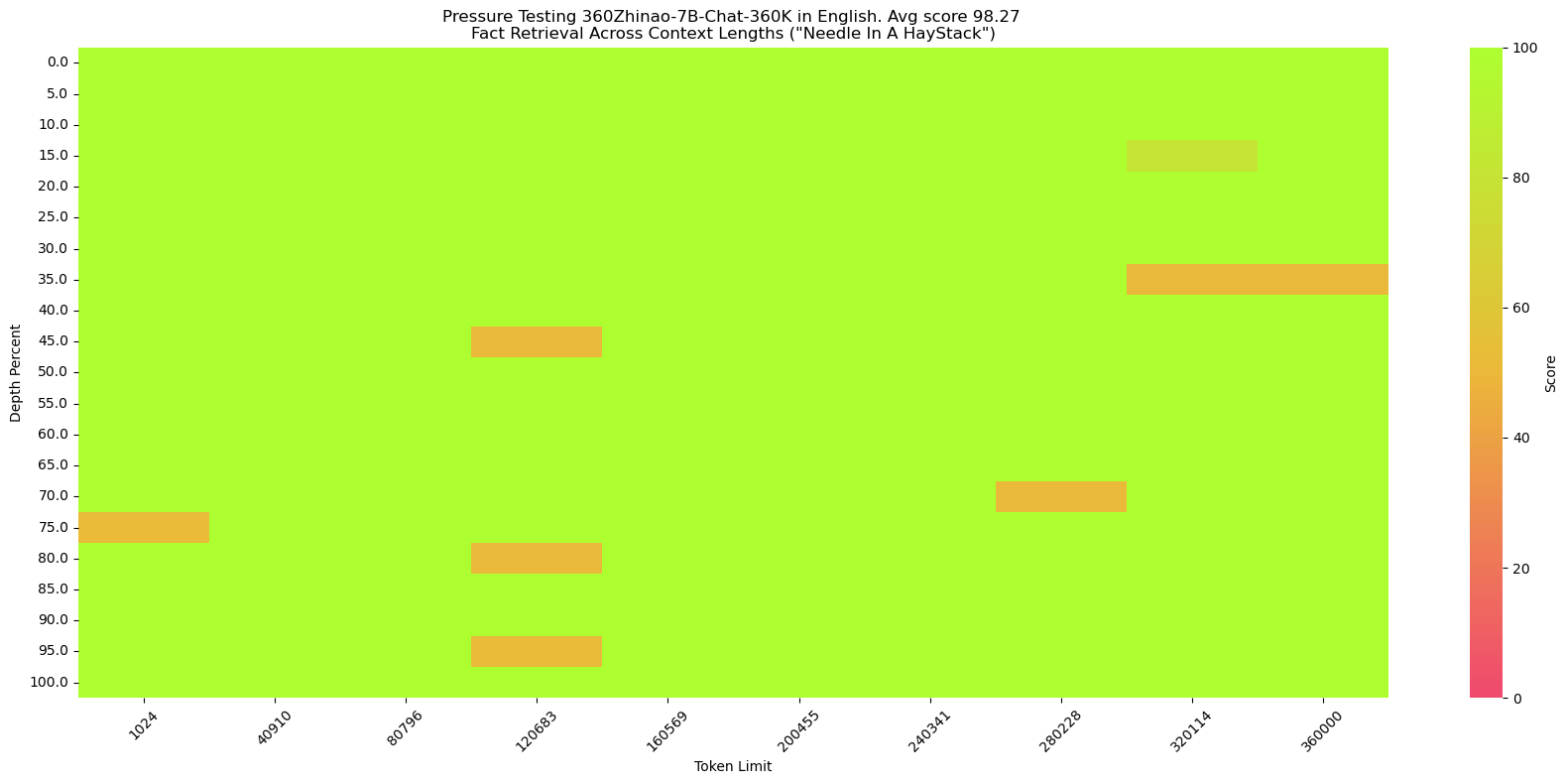}
    }
    \parbox{0.3\columnwidth}{\centering
    \includegraphics[width=0.3\columnwidth]{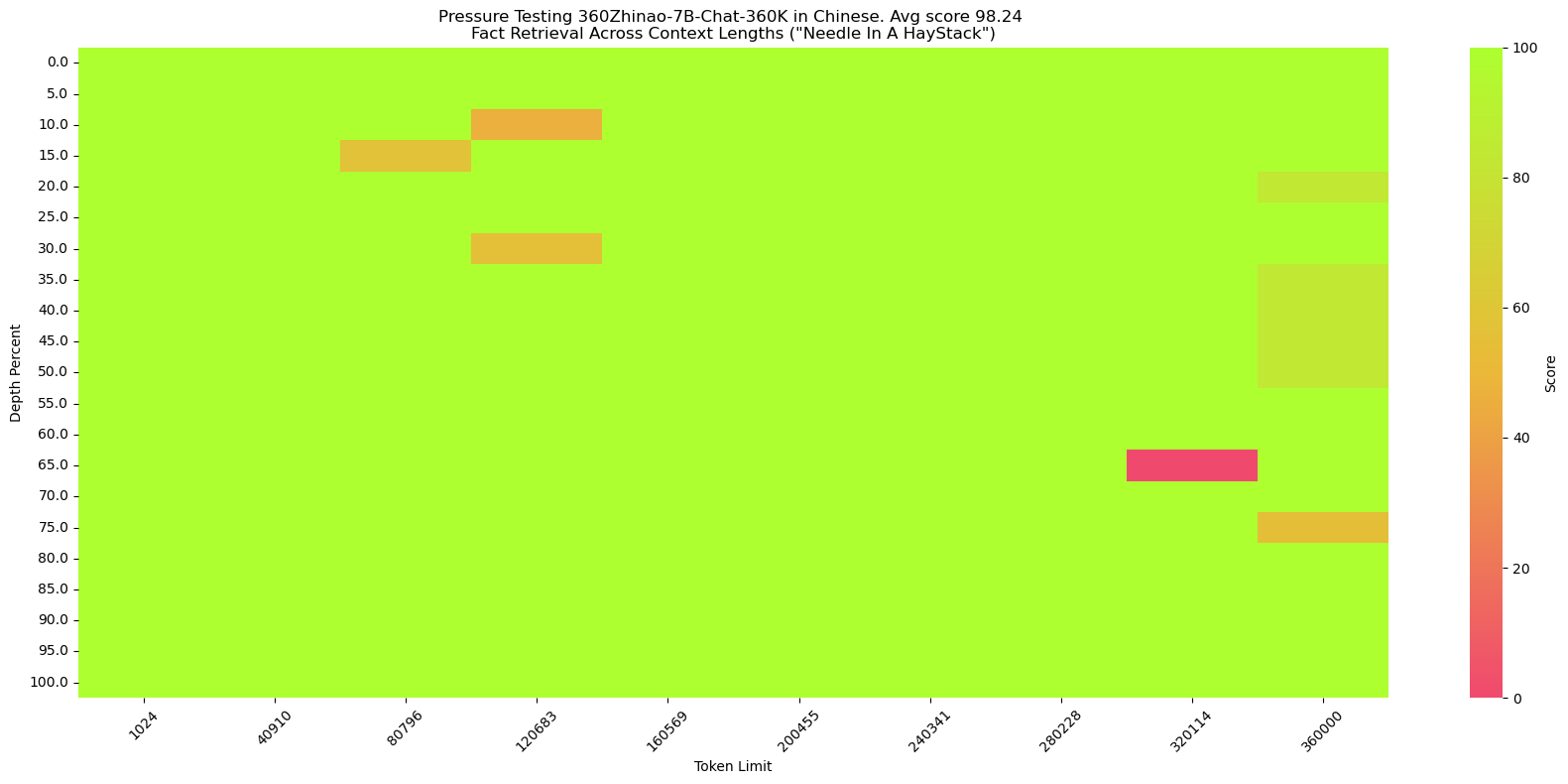}
    }
    \caption{From left to right: results on random city-number value retrieval \citep{niah2023v1}, original NIAH \citep{niah2023v0} and Chinese NIAH constructed by us. Value retrieval is relatively easy.}
    \label{fig:niah}
\end{figure}

We plot the results of the original version of NIAH and its adapted Chinese version in Figure \ref{fig:niah} (middle \& right). They are both much more difficult than the value retrieval variant. The details of the Chinese version could be found on our GitHub page. We have verified that our open-sourced 360Zhinao-7B-Chat-360K indeed achieves 100\% perfect all-green results on the value retrieval variant (Figure \ref{fig:niah}, left). The original versions are more tricky and the results are more unstable, depending on the inherent capability of different base models.

\textbf{Applications:}
When applying the 32K and 360K models in the backend of 360 AI-browser,
the 360k model, though end-to-end, exhibits considerable inference delay and is used offline to process long documents' results into cache. 
The 32K model is used online to process extremely-long documents in chunks in a parallel-recursive manner \citep{wu2021recursively} to interact with users in acceptable response time, giving user summaries, mind maps and QAs.

\subsection{RLHF}

\textbf{General RLHF:}
We have been working on RLHF since early 2023. Generally, we have continuously improved our PPO codebase and infrastructure and could support RLHF on large models of up to $\sim$100B size. 
We have accumulated over 1,500,000 high-quality pairs of preference data for RM training from filtered open-source data, feedback from our annotation team and past human comparison of multiple results on historical prompts.
The resulting RM essentially follows the RM scaling trends \citep{touvron2023llama,bai2022training,gao2023scaling} and achieves around 70\% - 75\% validation accuracy, increasing with the RM model size (internally 7B, 13B and 70B).
As for the RL algorithm we now mostly use a PPO variant similar to ReMax \citep{li2023remax}, discarding the value network in PPO.
On general LLM tasks as categorized by \citet{ouyang2022training}, PPO models usually achieves a little improvement on average, though not as significant as RLHF at OpenAI \citep{ouyang2022training}.

\textbf{Specific RLHF:}
PPO improvement on specific tasks are more often observed and has therefore become an indispensable step in domain-specific models.
The translation LLM behind our AI-browser improved a lot after PPO training in terms of translation quality and code switch (e.g. undesirable mixed Chinese-English output). The agent LLM after PPO exhibited more accurate function calling and lower false recall rate.
We posit this is because task-specific RMs are easier to train with more focused pairwise data, and rule-based rewards could be derived for code switch and function calling as the most accurate feedback reward to be added to the RM reward.

\textbf{RMs:}
An unexpected and nice by-product of our RLHF endurance is the versatility of RM. We found it quite pleasing to use our RM as judge to compare different versions and checkpoints, as referee to select the best response from multiple open-source competitive LLMs for each prompt and as filters to clean our dataset.
Deployed locally, our own RM accomplishes the aforementioned tasks in merely tens of seconds on a large dataset and gives satisfactory though not perfect results.

\textbf{DPO:}
Besides full RLHF, we have also observed recently positive signs of DPO variants, more specifically with ORPO \citep{hong2024reference} and NCA \citep{chen2024noise}. They demonstrated decent potential with LoRA \citep{hu2022lora} and we are working to reliably harness their gains.

\section{Conclusion, Limitation, and Future Work}

This technical report introduces our approaches to pretraining and finetuning 360Zhinao models.
During pretraining we explored various data cleaning and composition strategies and validated them with our carefully curated ablation environment that's designed to be stable and sensitive on minimal model size, which provides guidance to train our 7B model on 3.4T high-quality tokens.
When performing alignment we filtered and reformatted our SFT data to balance quantity over quality. We extended its supported sequence length to 32K and 360K with simple RoPE-base change and tailored long data.
Following SFT, RMs and RLHF are also trained and credibly applied to specific tasks and pipelines in our work.

Scope of this paper is limited to 7B models, and we hope to share our progress on larger models in the near future.
Training LLMs is an endless expedition and we are still constantly improving our data, infrastructure, model architecture and evaluation protocol.

\section{Author Contributions}

\begin{multicols}{2} 
Xiangzheng Zhang \\

\textbf{Pretraining:}

Jie Xiong \\
Zhuang Xiong \\
Cunqi Zhai \\
Lin Li \\
Huanyong Liu \\ 
Xiaochun Gong \\

\noindent
\textbf{Alignment:} \\
Qi An \\
Zhe Wei \\
Lifu Tang \\
Junchen Liu \\
Fenrui Xiao \\
Huan Liu \\
Liang Wen \\
Xin He \\
Haosheng Zou \\
Yongchao Deng \\
Shousheng Jia 

\end{multicols} 

We also thank all those who have contributed to 360Zhinao-7B models but are not mentioned in the paper, including but not limited to the bigdata team led by Guohui Liu, annotation team by Si Li, evaluation team by Fei Tan, and infra team by Xiaodong Sun.


\bibliography{iclr2024_conference}
\bibliographystyle{iclr2024_conference}

\appendix
\section{Data cleaning} \label{sec:app:clean}
The main rules in the data cleaning stage include:
\begin{enumerate}
    \item Removing documents with a high proportion of punctuation.
    \item Removing documents with a high proportion of paragraphs ending in ellipsis. 
    \item Remove documents with high paragraph ratios that lack punctuation endings; 
    \item Removing documents containing too many excessively long or short abnormal English words and numbers. 
    \item Removing documents with too many repetitive sentences; 
    \item Removing documents that contain too many short paragraphs; 
    \item Removing documents containing too many repeated n-grams (n=1,2,3). 
\end{enumerate}
Among the above rules, we set a relatively loose threshold for examination documents to avoid accidental deletion.

\section{Ablation result} \label{sec:ablation}

This section will report and analyze the ablation results of data strategies. We will plot training loss and validation loss graphs to aid in the analysis and present the ablation results of 360Eval, OpenCompass, and SFT in tabular form. To save space, the metrics in the table headers are abbreviated as follows: Bs (BrainStorm), Cot (Chain of Thought), Cls (Classification), Cqa (ClosedQA), Comp (Comprehension), Ext (Extraction), Gen (Generation), Sums (Summarization Short Text), Suml (Summarization Long Text), Oqa (OpenQA), Rewr (Rewrite), Tran (Translation), Mrc (Machine Reading Comprehension), Exam (Examination), Klg (Knowledge), Reas (Reasoning). The average of metrics of each evaluation is represented as Avg (Average) to indicate the model's final performance.

\subsection{Document deduplication} \label{sec:app:doc_dedup}

\begin{figure}[!htb]
  \centering
  \begin{subfigure}[b]{0.495\textwidth}
    \centering
    \includegraphics[width=\textwidth]{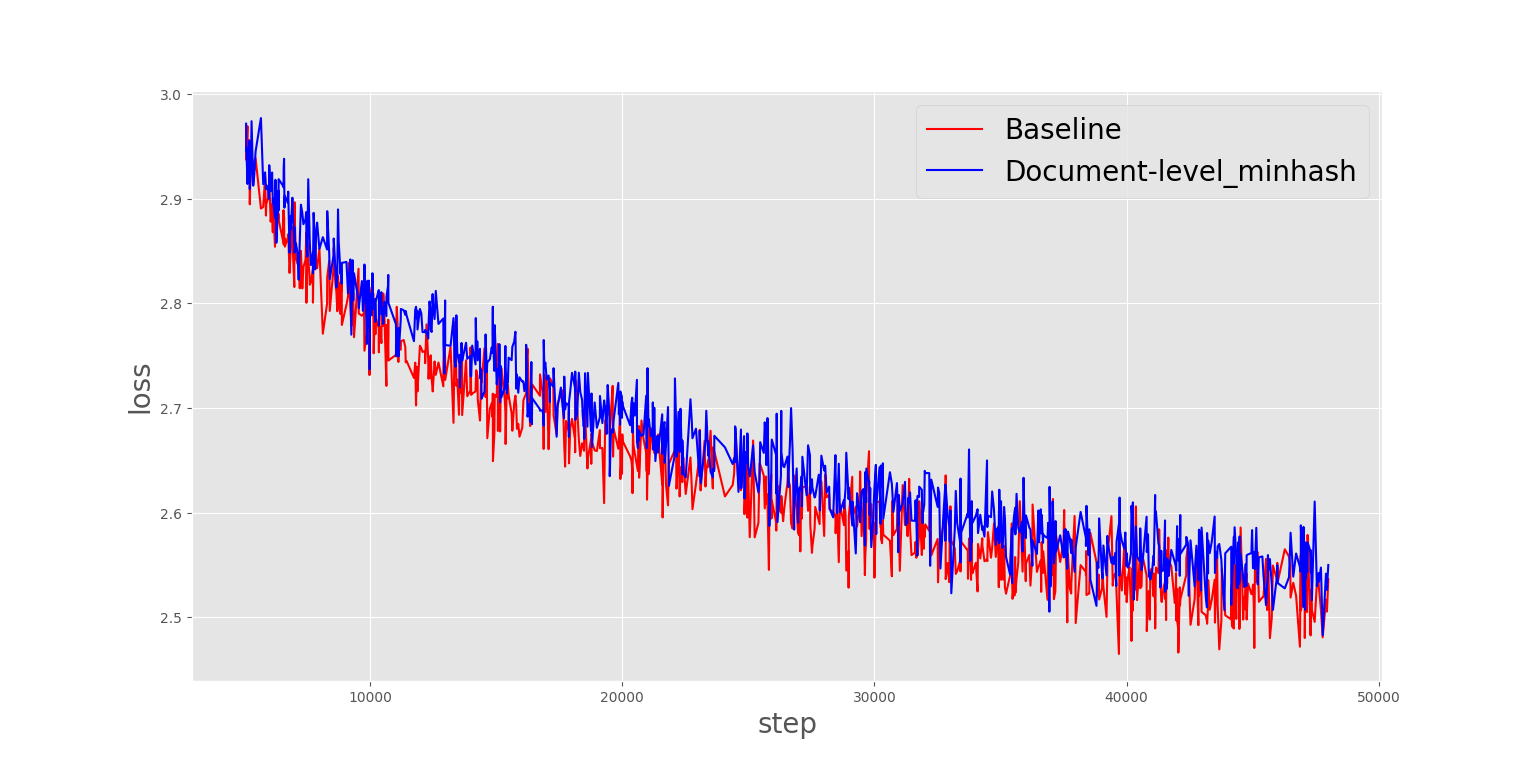}
    \caption{train loss}
    \label{fig:doc_dedup_1}
  \end{subfigure}
  \begin{subfigure}[b]{0.495\textwidth}
    \centering
    \includegraphics[width=\textwidth]{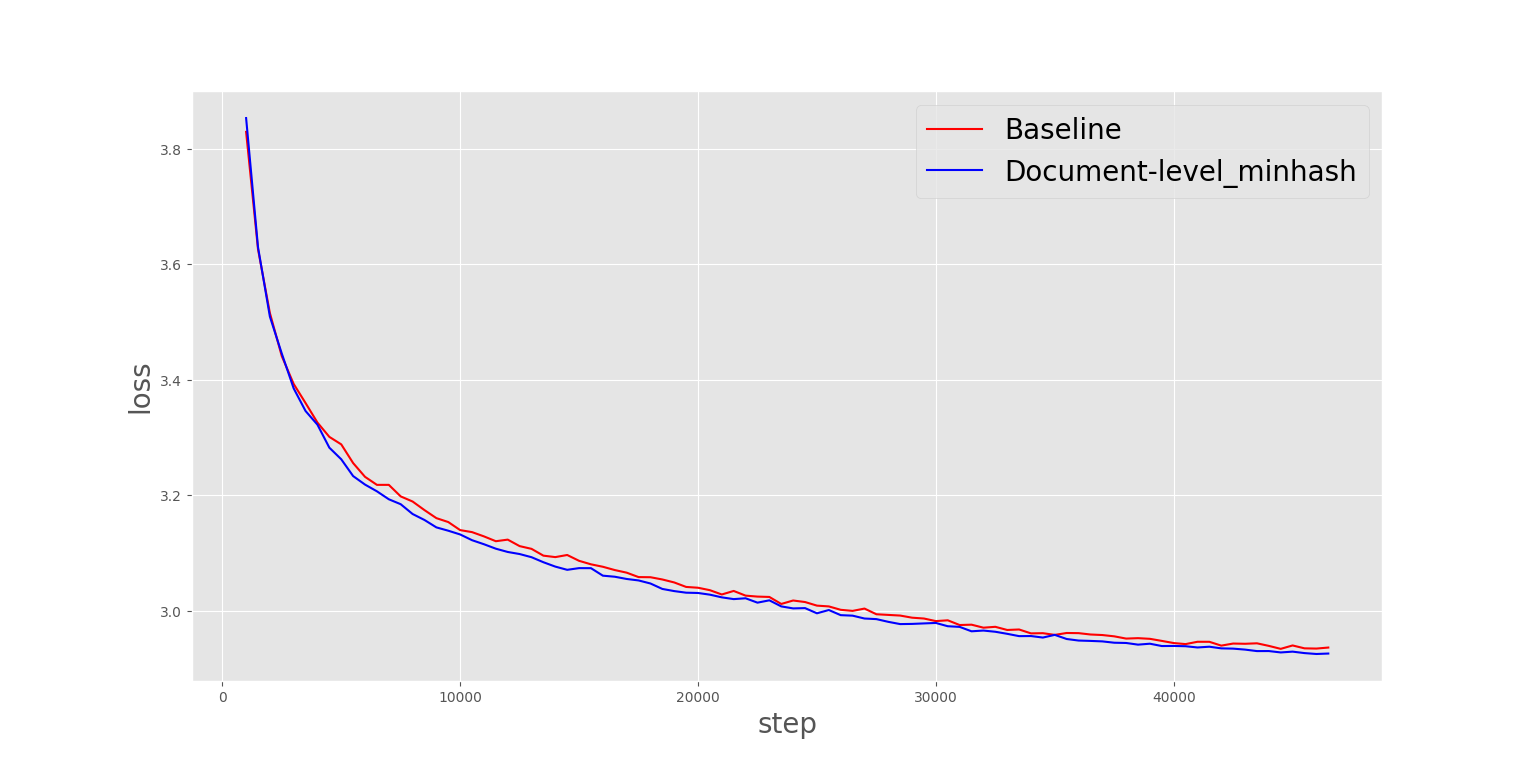}
    \caption{valid loss}
    \label{fig:doc_dedup_2}
  \end{subfigure}
  \caption{Loss curves of Document deduplication.}
  \label{fig:doc_dedup}
\end{figure}

\begin{table}[!htb]
\centering
\small
    \setlength\tabcolsep{3pt} 
    \renewcommand{\arraystretch}{1.25}
    \caption{Ablation results of document deduplication strategy on 360Eval and OpenCompass.}
\begin{tabu}{c|c|c|c|c|c|c|c|c|c|c|c|c|c} 
\hline
\multirow{2}{*}{Document} & \multicolumn{7}{c|}{360Eval}                         & \multicolumn{6}{c}{OpenCompass}              \\ 
\cline{2-14}
                          & \textbf{Avg}   & Mrc   & Ext   & Tran & Cls   & Exam  & Sum   & \textbf{Avg}   & Exam  & Lang  & Klg   & Usd   & Reas  \\ 
\hline
w/o dedup                 & 18.52 & 23.65 & 19.69 & 1.69 & 32.45 & 12.31 & 21.3 & 32.59 & 27.17 & 59.62 & 36.53 & 29.81 & 9.81   \\ 
\hline
w/ dedup                  & \textbf{18.77} & 24.33 & 19.3  & 1.93 & 37.6  & 8.98  & 20.5 & \textbf{32.84} & 25.54 & 63.46 & 36.22 & 28.83 & 10.13  \\
\hline
\end{tabu}
\label{table:doc_dedup_bench}
\end{table}

\begin{table}[!htb]
    \centering
    \small
    \setlength\tabcolsep{3pt} 
    \renewcommand{\arraystretch}{1.25}
    \caption{Ablation results of document deduplication strategy on SFT evaluation.}
    \begin{tabular}{c|c|c|c|c|c|c|c|c|c|c|c|c|c|c}
    \hline
        Document & \textbf{Avg} & Bs & Cot & Cls & Cqa & Comp & Ext & Gen & Sums & Oqa & Rewr & Suml & Tran & Code \\ \hline
        w/o dedup & 45.4  & 86.2 & 14.6 & 39.9 & 28   & 39.8 & 29   & 61.1 & 63.8 & 50.8 & 37.6 & 72.5   & 48.3 & 18.6  \\ 
        \hline
        w/ dedup  & \textbf{46.25} & 84.6 & 21   & 44.5 & 31.5 & 38.1 & 26.8 & 61.4 & 66   & 48.2 & 36.5 & 79.3   & 41   & 22.4  \\
    \hline
    \end{tabular}
    \label{table:doc_dedup_sft}
\end{table}
Figure \ref{fig:doc_dedup} displays the loss curves for document deduplication performed on web data. The loss curves show that the training loss of "document w/ dedup" is higher than that of "document w/o dedup", but the validation loss is lower. This suggests that the former exhibits better data diversity and results in a model with stronger fitting capability. Tables \ref{table:doc_dedup_bench} and \ref{table:doc_dedup_sft} present the results of document deduplication on benchmark and SFT, respectively. It can be observed that "document w/ dedup" performs better, with an improvement of 1.47\% on SFT, 0.99\% on 360Eval, and 0.02\% on OpenCompass. Our designed SFT and 360Eval evaluation systems are more sensitive to OpenCompass.

\subsection{Paragraph deduplication} \label{sec:app:para_dedup}

\begin{figure}[!htb]
  \centering
  \begin{subfigure}[b]{0.495\textwidth}
    \centering
    \includegraphics[width=\textwidth]{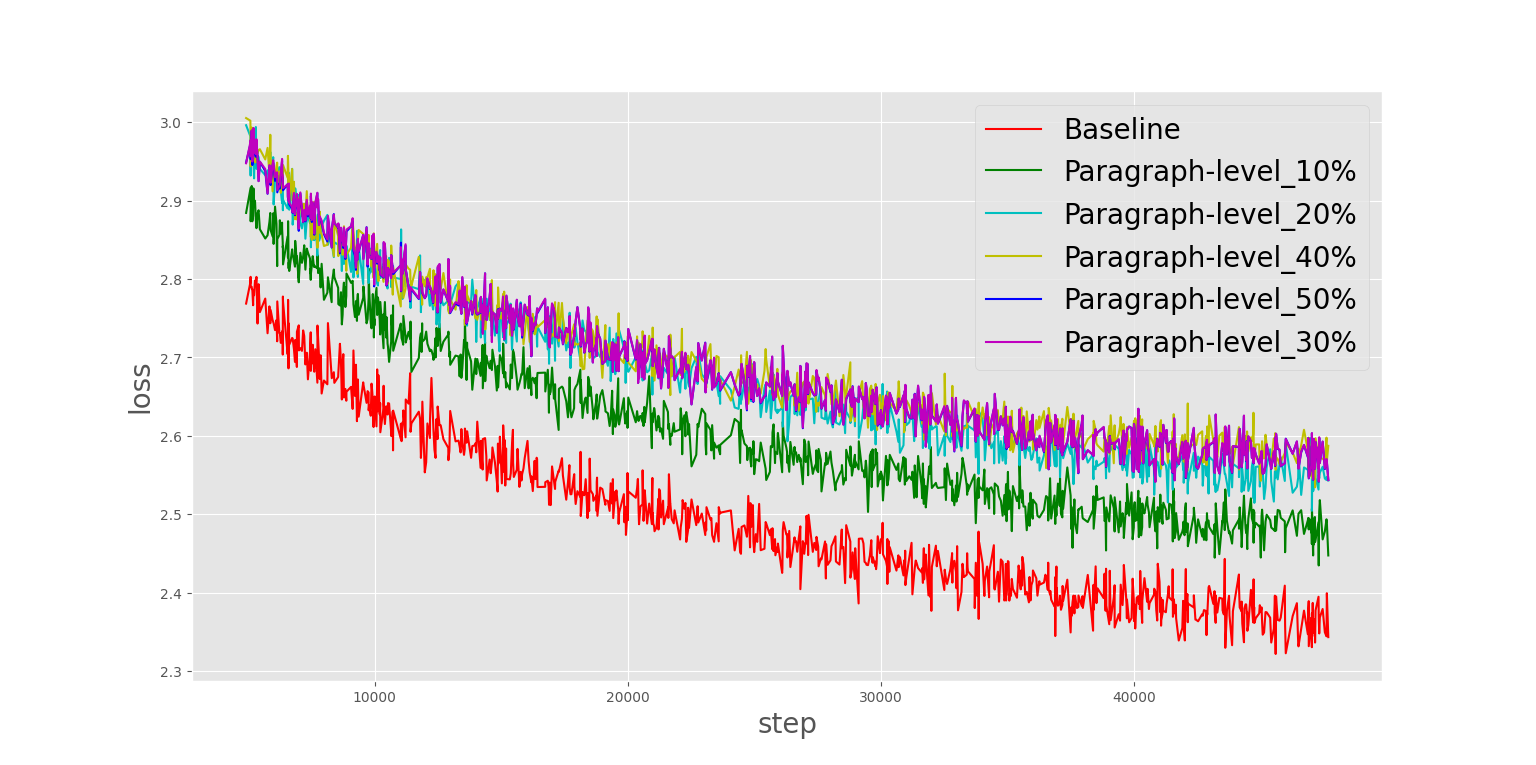}
    \caption{train loss}
    \label{fig:para_1}
  \end{subfigure}
  \begin{subfigure}[b]{0.495\textwidth}
    \centering
    \includegraphics[width=\textwidth]{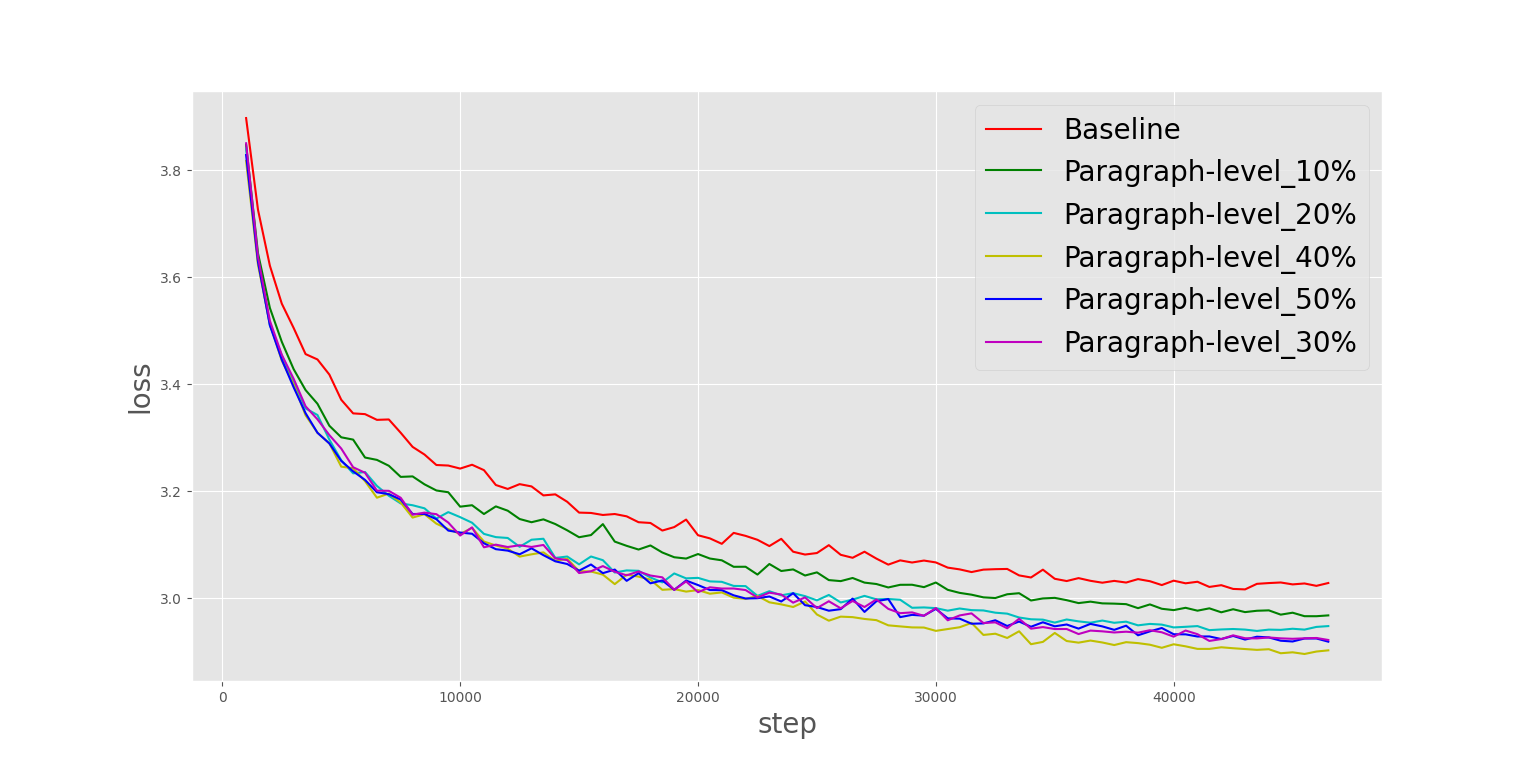}
    \caption{valid loss}
    \label{fig:para_2}
  \end{subfigure}
  \caption{Loss curves of Paragraph deduplication.}
  \label{fig:para_dedup}
\end{figure}

\begin{table}[!htb]
\centering
\small
    \setlength\tabcolsep{3pt} 
    \renewcommand{\arraystretch}{1.25}
    \caption{Ablation results of paragraph deduplication strategy on 360Eval and OpenCompass.}
\begin{tabu}{c|c|c|c|c|c|c|c|c|c|c|c|c|c} 
\hline
\multirow{2}{*}{Paragraph} & \multicolumn{7}{c|}{360Eval}                         & \multicolumn{6}{c}{OpenCompass}               \\ 
\cline{2-14}
                           & \textbf{Avg}   & Mrc   & Ext   & Tran & Cls   & Exam  & Sum   & \textbf{Avg}   & Exam  & Lang  & Klg   & Usd   & Reas   \\ 
\hline        
w/o~dedup               & 21    & 29.42 & 20.5  & 2.49 & 41.8  & 9.75  & 22.06 & 32.29 & 25.84 & 58.65 & 34.02 & 31.21 & 11.73  \\ 
\hline
dedup\_10\%                & 20.27 & 25.74 & 21.62 & 3.09 & 41    & 8.35  & 21.82 & 34.05 & 26.87 & 65.38 & 36.57 & 31.24 & 10.19  \\ 
\hline
dedup\_20\%                & 21.7  & 30.11 & 24.05 & 3.09 & 38.75 & 9.73  & 24.49 & 32.22 & 28.77 & 61.54 & 31.49 & 28.04 & 11.25  \\ 
\hline
dedup\_30\%                & \textbf{23.17} & 34.37 & 24.48 & 3.99 & 37.85 & 14.11 & 24.25 & \textbf{34.61} & 31.05 & 63.46 & 34.57 & 33.12 & 10.85  \\ 
\hline
dedup\_40\%                & 22.51 & 32.29 & 24.51 & 4.64 & 38.5  & 12.86 & 22.29 & 33.48 & 29.27 & 60.58 & 35.26 & 31.37 & 10.93  \\ 
\hline
dedup\_50\%                & 22.45 & 34.77 & 17.18 & 4.38 & 38.1  & 15.84 & 24.44 & 33.61 & 29.03 & 63.46 & 32.33 & 31.54 & 11.67  \\
\hline
\end{tabu}
\label{table:para_dedup_bench}
\end{table}

\begin{table}[!htb]
\centering
\small
\setlength\tabcolsep{3pt} 
\renewcommand{\arraystretch}{1.25}
\caption{Ablation results of paragraph deduplication strategy on SFT evaluation.}
\begin{tabu}{c|c|c|c|c|c|c|c|c|c|c|c|c|c|c} 
\hline
Paragraph   & Avg   & Bs   & Cot  & Cls  & Cqa  & Comp & Ext  & Gen  & Sums & Oqa  & Rewr & Suml & Tran & Code  \\ 
\hline
w/o~dedup & 47.92 & 86.3 & 19   & 39.4 & 30.2 & 42.6 & 36.2 & 61.3 & 66.1 & 50.4 & 44.9 & 76.3 & 48.8 & 18.4  \\ 
\hline
dedup\_10\%  & 48.79 & 84.3 & 19   & 46.4 & 33.3 & 45.9 & 34.3 & 61.7 & 67.4 & 48.9 & 40.2 & 70.8 & 53.5 & 28.6  \\ 
\hline
dedup\_20\%  & 49.75 & 86   & 15.9 & 47.4 & 36.8 & 40.7 & 32.1 & 63.7 & 73.2 & 51.7 & 40.3 & 80.4 & 51.8 & 26.8  \\ 
\hline
dedup\_30\%  & \textbf{53.04} & 87.5 & 28.4 & 43   & 36.8 & 51.2 & 39.1 & 65.3 & 75.8 & 50.2 & 52.8 & 79.2 & 52.9 & 27.3  \\ 
\hline
dedup\_40\%  & 51.41 & 88.9 & 17.3 & 43.5 & 33.7 & 51   & 43.2 & 67.7 & 69.8 & 54.8 & 44.4 & 79.6 & 55.9 & 18.6  \\ 
\hline
dedup\_50\%  & 52.33 & 86   & 26.7 & 52.3 & 36.2 & 49.9 & 38.6 & 65.2 & 75.9 & 50.9 & 47.1 & 81   & 48.1 & 22.4  \\
\hline
\end{tabu}
\label{table:para_dedup_sft}
\end{table}

Figure \ref{fig:para_dedup}, Table \ref{table:doc_dedup_bench}, and Table \ref{table:doc_dedup_sft} present the ablation results of paragraph deduplication at levels from 0\% to 50\%. From the loss curves, it is observed that the training loss of 30\%, 40\%, and 50\% paragraph deduplication is higher, but the validation set loss is lower, indicating that the paragraph deduplication strategy helps to improve data diversity. For ablation results, the best performance is achieved when the deduplication ratio is set to 30\%. Therefore, considering both data quantity and efficiency, we ultimately set the paragraph deduplication ratio to 30\%.

\subsection{Sentence deduplication} \label{sec:app:sent_dedup}
\begin{figure*}[!htb]
  \centering
  \begin{subfigure}[b]{0.495\textwidth}
    \centering
    \includegraphics[width=\textwidth]{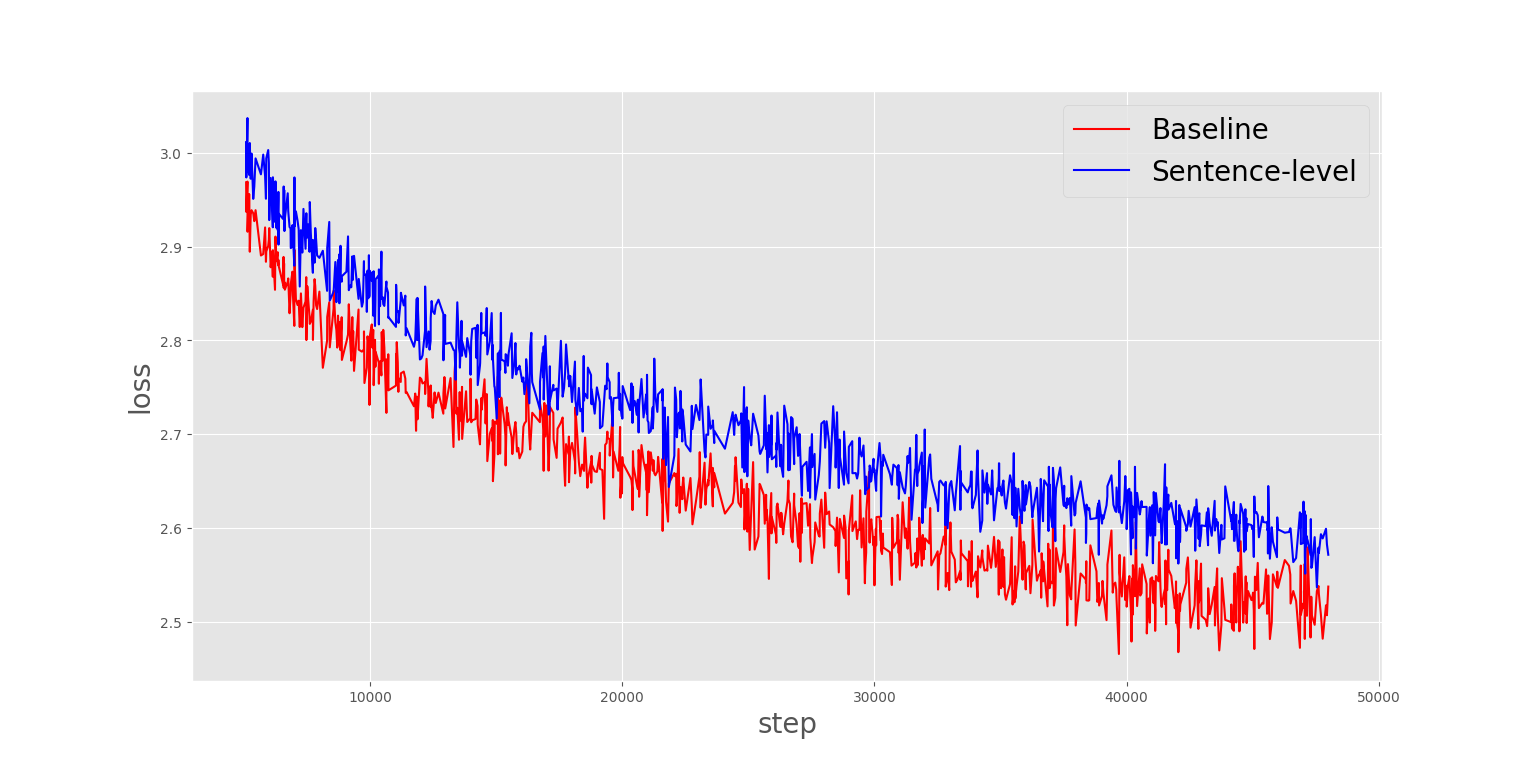}
    \caption{train loss}
    \label{fig:sent_dedup_1}
  \end{subfigure}
  \begin{subfigure}[b]{0.495\textwidth}
    \centering
    \includegraphics[width=\textwidth]{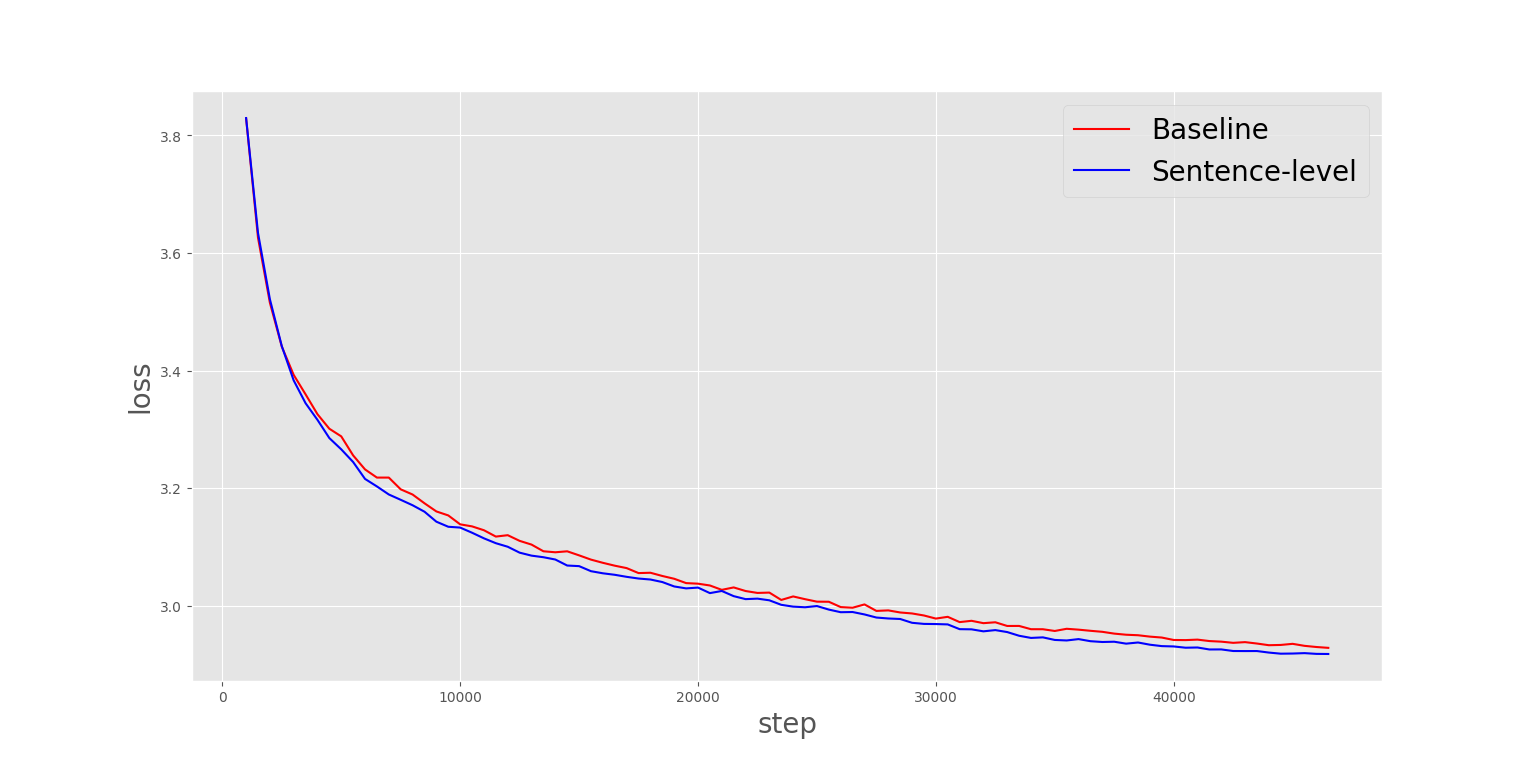}
    \caption{valid loss}
    \label{fig:sent_dedup_2}
  \end{subfigure}
  \caption{Loss curves of sentence deduplication.}
  \label{fig:sent_dedup}
\end{figure*}

Based on the findings from Figure \ref{fig:sent_dedup}, Table \ref{table:sent_dedup_bench}, and Table \ref{table:sent_dedup_sft}, it is evident that sentence deduplication significantly improves the performance of the model on 360Eval and SFT evaluation. Therefore, we integrate sentence deduplication as a data strategy into the recipe pipeline.

\begin{table}[!htb]
\centering

\begin{minipage}{\linewidth}
\centering
\small
\setlength\tabcolsep{3pt} 
\renewcommand{\arraystretch}{1.25}
\caption{Ablation results of sentence deduplication strategy on 360Eval and OpenCompass.}
\begin{tabu}{c|c|c|c|c|c|c|c|c|c|c|c|c|c} 
\hline
\multirow{2}{*}{Sentence} & \multicolumn{7}{c|}{360Eval}                        & \multicolumn{6}{c}{OpenCompass}              \\ 
\cline{2-14}
                          & \textbf{Avg}   & Mrc   & Ext   & Tran & Cls  & Exam  & Sum   & \textbf{Avg}   & Exam  & Lang  & Klg   & Usd   & Reas  \\ 
\hline
w/o dedup                 & 20.57 & 27.97 & 21.14 & 2.77 & 34   & 13.65 & 23.92 & \textbf{33.8}  & 28.94 & 64.42 & 36.53 & 30    & 9.14  \\ 
\hline
w/ dedup                  & \textbf{21.21} & 28.07 & 18.9  & 4.12 & 36.2 & 15.02 & 24.97 & 32.94 & 27.77 & 61.54 & 36.33 & 29.71 & 9.34  \\
\hline
\end{tabu}
\label{table:sent_dedup_bench}
\end{minipage}

\vspace{1cm} 

\begin{minipage}{\linewidth}
\centering
\small
\setlength\tabcolsep{3pt} 
\renewcommand{\arraystretch}{1.25}
\caption{Ablation results of sentence deduplication strategy on SFT evaluation.}
\begin{tabu}{c|c|c|c|c|c|c|c|c|c|c|c|c|c|c} 
\hline
Sentence  & \textbf{Avg}   & Bs   & Cot  & Cls  & Cqa  & Comp & Ext  & Gen  & Sums & Oqa  & Rewr & Suml & Tran & Code  \\ 
\hline
w/o dedup & 51.13 & 85.3 & 22.5 & 46.3 & 34.9 & 45.4 & 40.5 & 62.4 & 74.2 & 53.6 & 52.7 & 81.7   & 46.1 & 20.6  \\ 
\hline
w/ dedup  & \textbf{52.64} & 86.5 & 20.2 & 53.4 & 37.4 & 50.9 & 37.3 & 67.8 & 72.5 & 52.2 & 51.5 & 79.3   & 49.4 & 24.8  \\
\hline
\end{tabu}
\label{table:sent_dedup_sft}
\end{minipage}

\end{table}

\end{document}